\pdfoutput=1
\documentclass[journal]{IEEEtran}
\usepackage{cite}
\usepackage{amsmath,amssymb}
\usepackage{graphicx}
\usepackage{booktabs}
\usepackage{url}
\usepackage{xcolor}
\usepackage{multirow}
\usepackage[hidelinks]{hyperref}
\hypersetup{
  pdftitle={NI-ACT: A Neural Intent Interface for Brain-Embodied Manipulation},
  pdfauthor={Yinghao Li, WeiHsien Lee, Xing Li, Zeyu Zhang, Canyang Zhao, Ke Fang, Taowei Lu, Bing Liu},
  pdfkeywords={brain-computer interface, motor cortex, neural population, embodied policy, robotic grasping}
}

\newcommand{\method}{NI-ACT}
\newcommand{\hard}{Hard-7}
\newcommand{\soft}{Soft-2D}

\newcommand{\full}{Full-Latent}

\begin{document}
\title{NI-ACT: A Neural Intent Interface for\\Brain-Embodied Manipulation}
\author{Yinghao Li, WeiHsien Lee, Xing Li, Zeyu Zhang, Canyang Zhao, Ke Fang, Taowei Lu, and Bing Liu%
\thanks{This work was supported by the Brain Science and Brain-like Intelligence Technology-National Science and Technology Major Project, Grant 2026ZD0220500.}%
\thanks{Yinghao Li, Zeyu Zhang, Canyang Zhao, Ke Fang, Taowei Lu, and Bing Liu are with the Institute of Automation, Chinese Academy of Sciences, Beijing, China.}%
\thanks{WeiHsien Lee, Xing Li, and Bing Liu are with MindTrix.}%
\thanks{Zeyu Zhang is also with the University of Chinese Academy of Sciences, Beijing, China.}%
\thanks{Bing Liu is the corresponding author (e-mail: \texttt{bing.liu@ia.ac.cn}).}}
\maketitle
\thispagestyle{plain}
\pagestyle{plain}

\begin{abstract}
Intracortical brain--computer interfaces can drive robotic arms, but they keep
the user inside the fast sensorimotor loop: every approach, correction, and
grasp must be neurally specified. Whether intracortical population activity can
instead serve as a task-level condition for a learned embodied policy---without
symbolic commands or hand-designed skill libraries---remains unknown. \method{}
addresses this question with a hierarchical brain--embodied framework: a brief
motor-cortical signal is read once as a persistent intent, and a visuomotor
policy then closes the perception--action loop from vision and proprioception.
In a counterfactual grasping task where the scene cannot specify the target,
the neural intent alone determined which object the robot grasped; replacing
the intent changed the behavior accordingly. Passing the full population
latent as a dedicated intent token, rather than a one-hot decoded direction,
raised grasp completion from 69\% to 81\% over 132 held-out scenes, whereas a
continuous direction alone did not improve on a one-hot label. Where the signal
was read also mattered: in maze trials that dissociated the initial movement
from the final target, window-matched premotor (PMd) intent completed more
grasps than primary motor (M1) intent (77\% vs.\ 53\%). On straight reaches,
where movement and goal aligned, PMd and M1 decoded the goal similarly (83\%
vs.\ 81\%). This task-dependent pattern supports PMd as a higher-level intent
source that is especially useful when an embodied policy must act on a goal
independently of the initial movement.
Preparatory activity was itself sufficient: in a center-out task, a PMd+M1
intent read before the go cue---before any movement---drove complete grasps,
and in one recording it completed more grasps than the same animal's post-cue
readout. These results show that how a neural signal is represented, and where
and when it is read, jointly determine how effectively it conveys movement
goals to an embodied agent.
\end{abstract}


\section{Introduction}
Intracortical brain-computer interfaces (BCIs) have enabled people with paralysis to reach and grasp with robotic arms~\cite{Hochberg2012,Collinger2013,Wodlinger2015}. Most such systems decode kinematics continuously. The user specifies not only what to manipulate but also how to move, while every approach and correction depends on the neural channel. This places the brain inside the robot's fast sensorimotor loop and allows decoding noise to accumulate over a trajectory.

Motor control also exposes task goals before movement. Dorsal premotor cortex (PMd) represents candidate reaches during preparation, and preparatory and movement-related activity can occupy distinct population subspaces~\cite{Cisek2005,Churchland2010,Churchland2012,Kaufman2014,Elsayed2016}. When a barrier or distractor makes the initial hand trajectory differ from the eventual reach goal, PMd retains information about the goal while M1 more closely reflects the movement being executed~\cite{Cisek2005,Afshar2011}. A BCI might therefore read a goal once and delegate fast corrections to a controller that observes the robot and its environment. This division would change the role of the neural channel: it would specify the intended task, while perception and low-level movement remain in the robot's feedback loop.

This division also changes how errors matter. In continuous control, an erroneous velocity estimate can perturb one segment of a trajectory and may be corrected by subsequent neural commands. With a persistent task condition, an erroneous target can remain active for the entire episode, even while the robot executes each movement smoothly. The appeal of task-level control therefore depends on more than reducing the frequency of neural updates: the goal representation must be reliable, the policy must use it consistently, and the system must expose a way to revise an intent when the task changes. Our study isolates the first two requirements in a fixed-goal grasping task. It evaluates the sufficiency of a one-shot signal for a complete behavior, while leaving interactive intent revision and user workload for future human-in-the-loop tests.

Learned visuomotor policies offer an artificial counterpart of that lower level. Action-chunking transformers, diffusion policies, and recent vision-language-action models can locate objects, generate action sequences, and correct motion from camera and proprioceptive feedback~\cite{Zhao2023ACT,Chi2023DiffusionPolicy,Kim2025OpenVLA,Black2025Pi0,Liu2025RDT,Bjorck2025GR00T}. Their task condition is generally symbolic: a language instruction, a target image, or a one-hot label. A neural population state differs in kind from these conditions. It may carry graded structure shared across directions and trials, beyond the category selected by a decoder. Whether a policy can use that structure directly, without an intermediate symbol or a hand-designed skill library, has not been established.

This study asks whether a brief motor-cortical population state can itself be the task-level interface to an embodied policy. Three questions follow. First, does the neural intent determine the selected object, or can the policy infer it from the scene? Second, what representation is most useful to the policy: a decoded symbol, a continuous direction, or a population-space code? Third, where and when is a useful goal signal available, particularly when goal and movement are dissociated and before movement begins? These questions distinguish the availability of information in a neural recording from its usefulness to a downstream controller. Offline decoding accuracy alone need not predict closed-loop success: the policy must be able to consume the representation, not merely a decoder's ability to identify its label.

Task-level control therefore creates a distinct reliability tradeoff from continuous decoding. Because the neural condition is read once and held fixed, the policy can use feedback to correct its approach and contact but cannot ask the user to revise an intent that selected the wrong object. A valid evaluation must separate target selection from grasp completion and test whether changing only the neural input changes the selected target. It must also prevent the policy from reading the target directly from the scene. The counterfactual task makes each layout compatible with several target choices, and intent substitution tests the policy's dependence on the supplied neural condition. This establishes behavioral control of the robot by that input; it does not imply that the robot reproduces the neural circuit's own causal process.

\begin{figure*}[t]
\centering
\includegraphics[width=.98\textwidth]{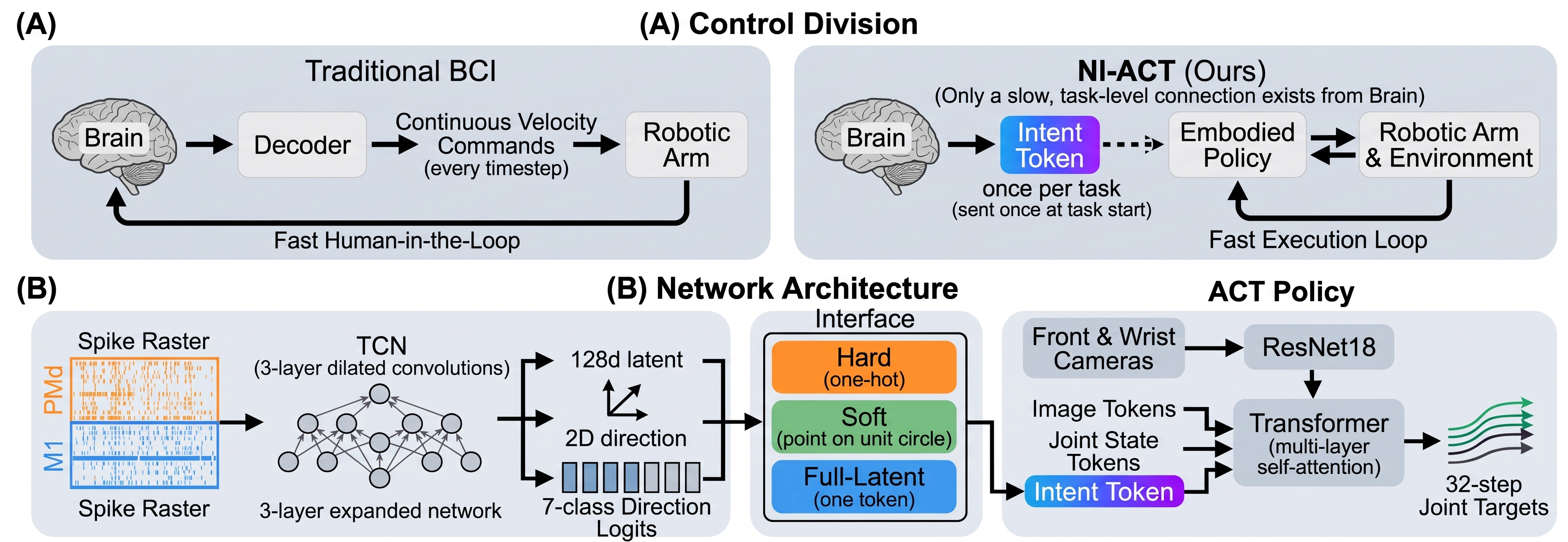}
\caption{NI-ACT links a brief motor-cortical readout to a learned visuomotor policy. Neural activity is encoded once as an intent; the policy then replans from camera images and joint state to produce robot actions. The experiments vary intent format, recording area, and readout time.}
\label{fig:framework}
\end{figure*}

We test these questions with NI-ACT, a hierarchical system in which a frozen neural encoder supplies one persistent intent and an action-chunking transformer (ACT) executes from camera images and joint state (Fig.~\ref{fig:framework}). We compare intent substitution, three intent formats, PMd and M1 sources, and pre-go versus later readouts. The results establish dependence of object choice on the transmitted intent, separate target selection from grasp execution, and identify preparatory motor-cortical activity as a useful condition for an embodied policy.

Neural calibration and policy training are intentionally separated. The encoder is fit to neural trials and their task labels, while the robot policy learns from images, joint states, and demonstrated actions; the two sources are joined only by the intended direction. This design tests whether a neural representation can function as a task condition across domains, rather than whether a model can memorize paired neural and kinematic trajectories. It also exposes a practical challenge: a neural encoder may retain trial-level or session-level variation that is irrelevant to robot execution, and the downstream policy must still learn a reliable mapping from its input to behavior.

The design makes distinct predictions for each question. If the signal is a cause of target choice, replacing it while leaving vision and the policy fixed should redirect behavior. If the population representation contains information useful beyond a decoded target class, a policy conditioned on that representation should complete more grasps after selecting the correct object. If PMd preferentially carries goal intent, its advantage should persist when PMd and M1 are read over matched time windows and when the animal's initial movement is dissociated from the goal. Finally, if preparatory activity already specifies the goal, a readout before the go cue should be sufficient to condition a complete grasp. The experiments test these predictions as properties of the neural-policy interface; they do not assume that the robot reproduces the animal's internal causal circuitry.

\section{Related Work}
\subsection{Neural Control and Goal Decoding}
Intracortical BCIs have demonstrated reach-and-grasp and control of robotic arms with many degrees of freedom in people with tetraplegia~\cite{Hochberg2012,Collinger2013,Wodlinger2015}. These systems typically decode continuous kinematics using Kalman-type or recalibrated decoders~\cite{Wu2006Kalman,Gilja2012Refit}. Continuous control preserves user authority at every step, but it also asks the brain to specify the trajectory itself: every approach and correction is expressed through the neural channel.

A complementary line of work decodes intended goals rather than movement. Cognitive prosthetics have decoded reach targets from posterior parietal cortex~\cite{Musallam2004,Aflalo2015}; shared-control frameworks infer user goals and arbitrate between user and robot~\cite{Dragan2013,Javdani2015}; and intracortical and non-invasive systems combine neural commands with vision-guided assistance~\cite{Downey2016,Muelling2017,Shared3D2023,ARShared2024}. NOIR maps decoded intent to parameterized skills~\cite{Zhang2023NOIR}, while AI copilots improve EEG-driven pick-and-place~\cite{AICopilot2025}. Together, these approaches establish that neural activity can specify what to do rather than how to move. Most, however, pass a decoded symbol or select from a predefined skill set. NI-ACT instead tests whether a learned embodied policy can be conditioned directly by a neural population representation.

Goal decoding and shared autonomy make the target explicit, then use a planner or selected skill to produce movement~\cite{Dragan2013,Javdani2015,Zhang2023NOIR}. These mappings are interpretable and useful when the goal set or action repertoire is fixed. NI-ACT asks whether one learned visuomotor policy can interpret a neural condition through visual demonstrations, without a hand-specified mapping from each decoded class to an action primitive. It evaluates one-shot intent for fixed-goal grasping; continuous control and user correction remain relevant when goals change.

The gap between decoder accuracy and behavior is central to closed-loop BCI evaluation. Held-out decoding measures prediction of neural labels, whereas online performance also depends on the device, task, and feedback loop~\cite{Glaser2020Decoding,Orsborn2014}. We separate target selection from completed grasps and substitute the neural input while holding the scene and policy fixed. Because recordings and demonstrations are independent, these tests assess the neural-policy interface rather than a person controlling the robot in real time.

\subsection{Motor-Cortical Population Representations}
The timing and organization of motor-cortical activity motivate a task-level readout. PMd represents candidate reaches before selection~\cite{Cisek2005}; preparatory activity shapes movement initiation~\cite{Afshar2011}; and preparation and execution can occupy distinct population subspaces~\cite{Kaufman2014,Elsayed2016}. PMd also carries rich task-relevant structure, with hand position, movement direction, speed, and reward location co-encoded in the same population~\cite{Cao2025PMd}. Maze tasks with barriers or distractors separate the final goal from the initial hand direction, making this distinction measurable.

Motor-cortical goals are population phenomena: latent-variable models recover structured single-trial dynamics~\cite{Pandarinath2018LFADS,Pei2021NLB,Gallego2020}, and brief windows support goal decoding~\cite{Pei2021NLB}. A decoded label or kinematic readout compresses that state, but offline accuracy does not show which representation a robot policy can use. We compare a class, a continuous direction, and a population code in closed-loop execution, and test PMd versus M1 and pre-go versus later readouts. The learned latent is treated as an empirical policy condition, not as a direct estimate of a biological preparatory manifold.

\subsection{Learned Visuomotor Policies}
Action-chunking transformers~\cite{Zhao2023ACT}, diffusion policies~\cite{Chi2023DiffusionPolicy}, and vision-language-action models~\cite{Kim2025OpenVLA,Black2025Pi0,Liu2025RDT,Bjorck2025GR00T} map observations and a task condition to action sequences. These models provide an effective low-level controller because they use camera and state feedback to replan during execution. Their conditioning interfaces are generally language, target images, or discrete labels. NI-ACT changes the condition itself: a cortical signal is read once, then consumed by one learned policy while that policy handles perception and movement. The central empirical issue is not whether an intent can be decoded offline, but whether its representation supports reliable policy execution.

In behavior-cloned manipulation, the observation and demonstrated action can carry correlated clues about the goal. A policy may therefore succeed by exploiting scene regularities even when its task condition contributes little. This makes a counterfactual evaluation important: the same visual layout must support different valid actions, with the intent distinguishing which one to perform. Our demonstrations vary the target within each layout, and our rollout intervention changes the neural input while holding the scene and policy fixed. The outcome is evaluated as a complete closed-loop grasp, so the comparison tests how a learned controller uses the condition during embodied execution rather than only how well an encoder predicts a label.

\section{Framework and Experimental Design}
\subsection{Neural Intent and Embodied Policy}
For spike-count window $S\in\mathbb{R}^{T\times N}$, the frozen neural encoder $E_\theta$ returns a population latent $z\in\mathbb{R}^{128}$, a two-dimensional direction $\hat{v}$, and seven-way direction logits $\ell$. An interface $G$ selects the condition passed to the policy:
\begin{equation}
 \begin{aligned}
 (z,\hat{v},\ell)&=E_\theta(S),\qquad u=G(z,\hat{v},\ell),\\
 \hat{A}_{t:t+31}&=\pi_\phi(I_t,q_t,u).
 \end{aligned}
\end{equation}
Here, $S$ comprises spike counts binned at 20-ms intervals within the task-specific readout window, while $I_t$ denotes the front and wrist images and $q_t\in\mathbb{R}^{9}$ the robot joint state. The three interfaces are \hard{} (one-hot argmax of $\ell$), \soft{} ($\hat v$), and \full{} ($z$). Each maps the same encoder output to a different policy condition: a discrete target class, a continuous two-dimensional direction, or the complete population latent. The neural window is read once before the grasp episode and $u$ remains fixed; at each robot update, the policy receives new images and joint state and predicts the next action chunk. This separates task specification from movement execution. The policy receives no target coordinates or object labels, so vision identifies the objects but cannot reveal which one is intended. All interface comparisons use the same frozen encoder, demonstrations, policy architecture, and evaluation scenes.

\begin{figure*}[t]
\centering
\includegraphics[width=.96\textwidth]{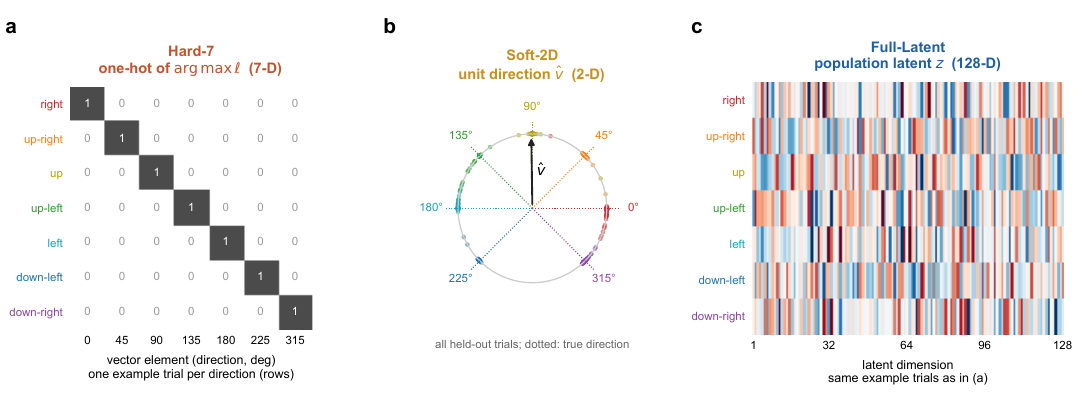}
\caption{The three neural intent representations from one frozen encoder (monkey J, PMd, movement-onset window; 426 held-out trials). (a) Hard-7 is the one-hot argmax of the seven-class head. (b) Soft-2D is the continuous unit direction for each held-out trial. (c) Full-Latent shows the 128-D population code for example trials, where each direction gives a distinct pattern across latent dimensions.}
\label{fig:interfaces}
\end{figure*}

The interfaces preserve progressively different aspects of the encoder output (Fig.~\ref{fig:interfaces}). Hard-7 retains only the winning class; Soft-2D retains a continuous reach direction; Full-Latent passes the complete projected population state. Thus the comparison tests whether the policy benefits from information that is discarded when the neural state is reduced to a direction or label.

The encoder is a temporal convolutional network with three one-dimensional convolutional blocks (256 channels, kernel size 5, dilations 1, 2, and 4), followed by temporal mean and max pooling and a projection to the 128-dimensional latent. Two supervised heads read out direction logits and a continuous unit vector. Training minimizes seven-way cross-entropy plus a 0.2-weighted squared error on the two-dimensional direction, with 10\% spike dropout and early stopping on validation trials. The encoder is then frozen. Consequently, policy comparisons cannot change the neural features to favor one interface; the same calibrated representation is passed to each policy for a given neural source.

The ACT policy combines the intent with two camera views and the nine-dimensional robot state. A shared ImageNet-pretrained ResNet-18 converts each $224\times224$ image to a $7\times7$ feature map, or 49 spatial tokens per camera. A four-layer transformer encoder of width 512 and eight attention heads processes those tokens together with the state and latent tokens. The one-layer decoder uses 32 learned action queries and a linear output head to predict 32 future nine-dimensional joint targets. Hard-7 and Soft-2D are concatenated with the joint state, producing 16- and 11-dimensional state inputs, respectively. Full-Latent is normalized separately, projected from 128 to 512 dimensions, and supplied as a dedicated intent token. The three policies share the visual and transformer architecture but have separate weights, allowing each to learn its own mapping from condition to action.

Each policy is trained by behavior cloning for 20,000 steps with batch size 64, using learning rates of $10^{-4}$ for the transformer and $10^{-5}$ for the image backbone. In standard ACT, a conditional VAE encodes a demonstrated action chunk and robot state into an action-style latent; the policy uses this variable with the current observations to generate a chunk, allowing the model to represent variation across demonstrations~\cite{Zhao2023ACT}. The posterior encoder is discarded at test time, when ACT sets this latent to the mean of its unit-Gaussian prior (zero) for deterministic decoding. In our experiments, we disable the CVAE pathway and hold its separate 32-dimensional input slot at zero during both policy training and evaluation. This constant ACT latent carries no trial-specific action mode or target information. It is distinct from the 128-dimensional neural latent $z$, which is the Full-Latent intent condition. Fixing the ACT latent keeps the neural intent as the only target-specific policy input in the counterfactual task, while visual feedback guides approach and contact; this controlled choice does not imply that CVAE-based ACT is less suitable for other tasks.

The three representations are different bottlenecks on the same learned state. Hard-7 discards confidence and all within-class structure; Soft-2D retains a graded spatial direction; Full-Latent exposes every latent coordinate to the policy. Full-Latent therefore bypasses a discrete readout at deployment, but its encoder still receives direction supervision during calibration. It should not be interpreted as an unsupervised or label-free interface. Separate policies are trained for each condition so that each learns to use its own input format; the comparison holds the architecture and training budget fixed, rather than the fitted policy weights.

Two controls test alternative explanations for a latent advantage. The class-prototype condition replaces each trial latent with the mean training latent assigned to its decoded class, retaining a 128-D class code while removing trial-specific variation. The Hard-7-token condition sends the seven-dimensional one-hot through the dedicated-token pathway used by Full-Latent. These controls distinguish the need for trial variation from the choice of input pathway, although they do not fully separate dimensionality from class geometry. In all cases, the encoder is frozen before behavior cloning: robot action errors update the policy, not the neural representation.

\begin{figure*}[t]
\centering
\includegraphics[width=.98\textwidth]{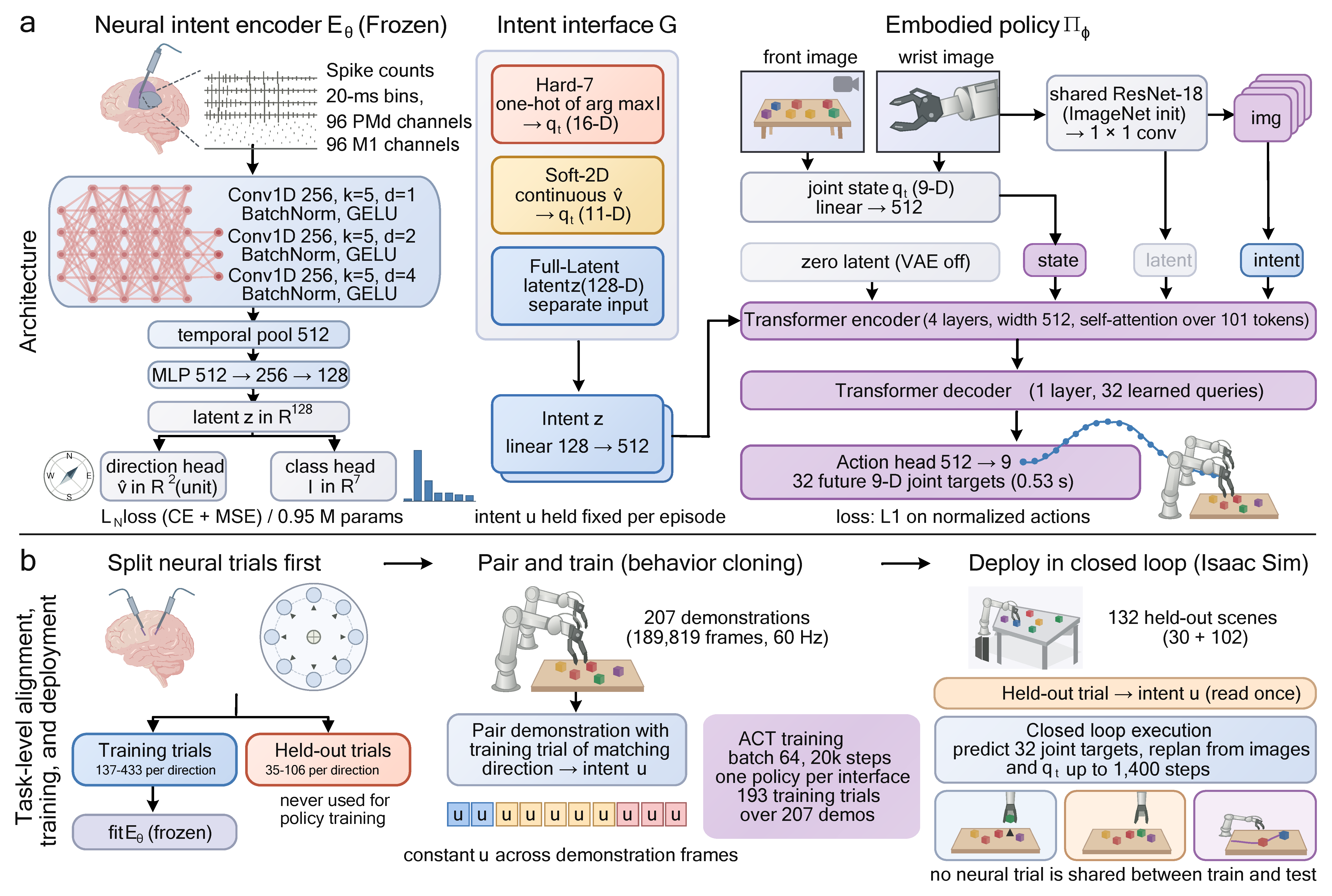}
\caption{Architecture and data protocol of NI-ACT. (a) The frozen temporal convolutional encoder maps binned spikes to a 128-D neural population latent, a 2-D direction, and seven-class logits. The dual-array illustration shows 96 PMd and 96 M1 channels; the primary MC\_Maze closed-loop encoder used 182 units. Hard-7 and Soft-2D are appended to the joint-state token; Full-Latent is projected from 128 to 512 dimensions as a separate neural-intent token. The ACT action-style latent is a different 32-D variable, fixed at zero with the CVAE pathway disabled. The transformer also receives 98 image tokens (49 per camera) and predicts 32 future joint targets. For Hard-7 and Soft-2D the encoder processes 100 tokens; for Full-Latent it processes 101. (b) Neural trials are split before pairing with the 207 counterfactual demonstrations. Each demonstration receives an intent from a training trial with the same direction; at deployment, each of 132 held-out scenes receives a held-out neural trial. Intent stays fixed during each episode while the policy replans from visual and proprioceptive feedback. No neural trial is shared between policy training and evaluation.}
\label{fig:method}
\end{figure*}

Figure~\ref{fig:method} summarizes the full conditioning pathway and evaluation protocol. The encoder supplies task-related signals, the interface determines which information is preserved, and ACT learns how that condition maps to actions while repeatedly observing the scene. Each interface uses the same demonstrations, train/test partition, policy objective, and frozen encoder. This separates neural calibration from robot learning: the intent indicates which target to pursue, while images and joint state guide approach, alignment, and grasp execution. Accordingly, we report both target selection and completed grasps rather than treating a correct decoded direction as successful control.

The system operates on two timescales. A neural window is encoded once before each episode, and the resulting condition remains fixed; the policy predicts 32 joint targets, executes that chunk, then replans from updated camera and joint-state observations (about 0.53 s per chunk). Thus the neural channel sets the task while the robot feedback loop handles the approach and contact sequence. A mistaken target and an incomplete grasp therefore have distinct signatures: the former concerns intent selection, while the latter concerns policy execution after selection. This study evaluates fixed-goal episodes, so it does not test how a user might update intent if a goal changes mid-episode. The paired-scene substitution isolates the role of the input at the task level: the same trained policy and visual state receive either the recorded intent or a direction-mismatched intent.

\subsection{Counterfactual Task and Evaluation}
The primary closed-loop study uses MC\_Maze neural recordings (DANDI 000128)~\cite{Pei2021NLB} and simulated Franka-arm demonstrations. Additional neural datasets and readout configurations are summarized in Table~\ref{tab:datasets}.

\begin{table*}[t]
\centering
\caption{Neural datasets and held-out splits.}
\label{tab:datasets}
\footnotesize
\setlength{\tabcolsep}{4pt}
\begin{tabular}{@{}llccc@{}}
\toprule
Animal & Dataset & Input & Split & Readout \\
\midrule
J & MC\_Maze (000128) & 182 units & 1505/300/449 & Go-cue aligned \\
J & Dual-array (000070) & 96/96 ch. & 1433/286/426 & Pre/post-go, move \\
M & Session A (2015-06-10) & 31/18 units & 164/41 & Post-go \\
M & Session B (2015-05-12) & 60/30 units & 408/103 & Pre/post-go \\
T & Five sessions (000688) & 64 PMd ch. & 754/192 & Prep./exec. \\
\bottomrule
\end{tabular}
\end{table*}

The datasets serve distinct tests and are calibrated separately. Monkey J's MC\_Maze recording supplies the primary interface comparison. Its one-second neural window extends from 560 ms before to 440 ms after the go cue; “go cue” in Table~\ref{tab:datasets} names the alignment event ($t=0$), while these bounds give the window duration and position. The dual-array maze recording supplies the area comparison: its paired channel counts are listed PMd/M1, and barriers or distractors make the initial hand movement differ from the eventual target in 79\% of trials. Monkey M's center-out sessions test the framework in another task and include a pre-go readout, while monkey T's five PMd sessions characterize goal information across preparatory and execution windows. Table splits are train/validation/test for monkey J and train/test for monkeys M and T; the paired PMd/M1 counts are in that order. Monkey M's raw sessions contained 208 and 563 trials, of which 205 and 511 had finite go-cue times and enter the listed splits. Monkey T's row pools independent splits from its five earliest center-out sessions (2013-08-19, 2013-08-21, 2013-08-23, 2013-09-03, and 2013-09-05). Each animal's encoder and policy were calibrated separately.

Robot demonstrations and neural recordings are independent. Each training demonstration is paired with a random neural training trial of the same direction, and each evaluation scene receives a held-out neural trial of its target direction. For MC\_Maze, each direction contributes 137--433 training and 35--106 held-out neural trials. The 207 demonstrations use 193 distinct training trials, with no trial used more than twice; the 132 evaluation scenes use 108 distinct held-out trials, with no trial used more than three times. Splitting neural trials before pairing prevents a neural sample used to condition behavior-cloning updates from reappearing during evaluation. The robot policy learns only from its own image, joint-state, and action sequences; the direction label is the link between the neural and robot datasets.

Three colored cubes were placed at three of seven directions around the workspace center with pairwise separation of at least $90^\circ$; color assignment and spatial jitter varied the layouts. The same layout was demonstrated once for each cube as target, producing three valid scene-action pairings and making vision alone insufficient to infer the goal. Training used 207 successful demonstrations from 69 layouts (189,819 frames). The held-out evaluation combined 132 scenes collected after training: 30 primary scenes from 10 layouts and 102 additional scenes from 34 new layouts. Test targets included cases whose scripted demonstration had failed, so evaluation was not restricted to teacher-success examples. We pooled the two scene groups for the three-interface comparison; analyses run only on one group are identified as such.

Every rollout began from the robot's home pose and ran for up to 1,400 control steps. After each 32-step action chunk, the policy received updated camera images and joint state and replanned; the neural intent remained unchanged throughout the episode. We scored target selection as the cube nearest the hand at gripper closure, and counted a completed grasp when the intended cube was lifted at least 10 cm. A lift of a non-target cube was recorded as a wrong-object grasp; no lifted cube was a failed completion. This separation distinguishes an intent that selects the wrong object from a policy that selects correctly but fails during execution. For the substitution control, a trial from another direction replaced the original neural input while the scene, policy, and execution protocol stayed fixed.

\section{Results}
\begin{figure*}[t]
\centering
\includegraphics[width=.98\textwidth]{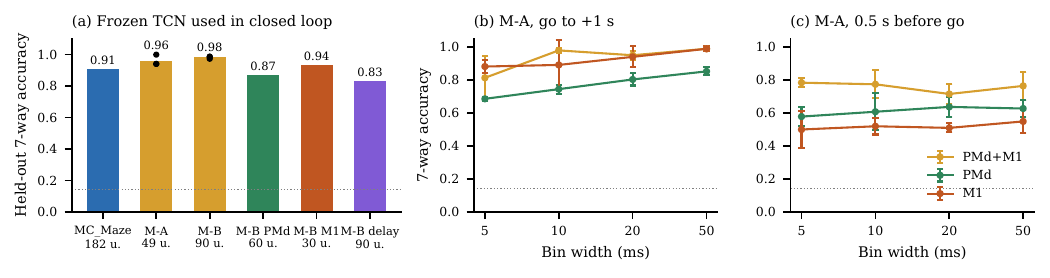}
\caption{Held-out goal-decoding accuracy across neural conditions. The MC\_Maze encoder used for closed-loop experiments reached 90.6\% on 449 held-out trials. Monkey M center-out sessions reached 94--100\% in session A and 97--99\% in session B across encoder seeds; accuracy was stable across 5--50 ms bins. In session B, the 500-ms delay-period readout was above chance before the go cue.}
\label{fig:decoding}
\end{figure*}

The first requirement is that the goal be recoverable from the recordings at all. The frozen encoders read held-out goal direction well above the seven-way chance level (14.3\%): the MC\_Maze encoder used for the primary study reached 90.6\% on 449 held-out trials, and the monkey-M center-out sessions reached 94--100\% (session A) and 97--99\% (session B) across seeds (Fig.~\ref{fig:decoding}). Offline decoding is therefore a solved prerequisite here. The experiments below address the open question: whether a downstream policy can turn one readout into a completed grasp, and which readout lets it.

\subsection{The Neural Intent, Not the Scene, Selects the Object}
A task-level interface is causal only if the neural condition controls behavior: change it while holding scene and policy fixed, and the robot must change what it does. With decoded MC\_Maze intent, the Full-Latent policy approached the intended object in all 30 primary scenes, completed 27 grasps (90\%), and never lifted a wrong object. Substituting a held-out trial from a different direction reduced grasps of the original target to zero and selection of that target to 1--3 of 30 scenes across the three interfaces.

\begin{figure*}[t]
\centering
\includegraphics[width=.98\textwidth]{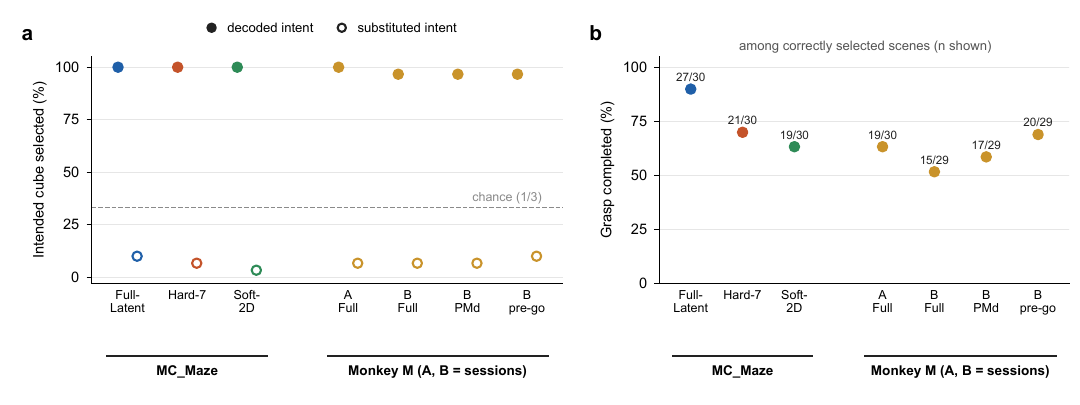}
\caption{Target selection and grasp completion. (a) Fraction of scenes in which the robot approached the intended object with decoded or substituted intent. (b) Grasp completion among correctly selected targets; labels show completed/selected counts. The panel includes the MC\_Maze interface conditions and additional monkey-M session, PMd-only, and pre-go evaluations.}
\label{fig:selection}
\end{figure*}

Because only three of seven directions held a cube, a substituted intent could point at empty space rather than another object. Of the 30 Full-Latent substitutions, 14 pointed at empty locations and could not test a switch; among the 16 that pointed at another cube, the robot followed the substituted direction in 13 and lifted nothing in the remaining three, never grasping the original target (Fig.~\ref{fig:selection}). The substitution thus redirected object choice with no change to the visual scene.

The original-intent results additionally separate selection from execution: every target was selected correctly, yet only 27 of 30 grasps completed. Selection is thus under neural control, whereas completion is a property of the policy; the remainder of the section turns to that second stage.

\begin{figure*}[t]
\centering
\includegraphics[width=.98\textwidth]{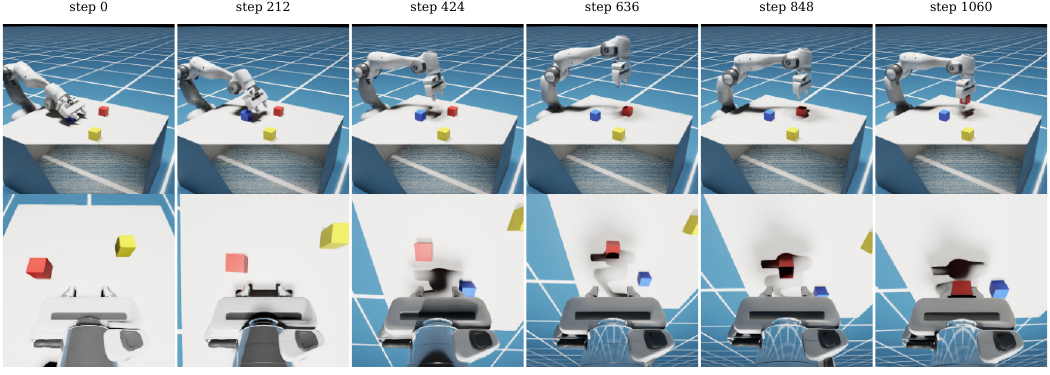}
\caption{A held-out Full-Latent rollout with the red cube as target. Front-camera views (top) and wrist-camera views (bottom) show approach and grasp execution. One neural intent remains fixed throughout; subsequent action chunks use updated camera and joint-state feedback.}
\label{fig:rollout}
\end{figure*}

Figure~\ref{fig:rollout} illustrates this division of labor in one held-out episode: the intent is fixed once, and successive camera observations alone support approach, descent, closure, and lift, with no phase-specific neural command.

\subsection{The Population Code, Not a Decoded Symbol, Best Supports Grasping}
Given that the intent selects the object, the next question is which representation best converts a selected target into a completed grasp. The three interfaces compress the same frozen encoder output to progressively less information: Full-Latent keeps the 128-dimensional population code, Soft-2D keeps only a continuous direction, and Hard-7 keeps only the winning class. The comparison therefore asks which neural condition gives behavior cloning the most useful information for approach, alignment, and lift.

Full-Latent completed 107/132 intended grasps (81.1\%), versus 91/132 (68.9\%) for Hard-7 and 86/132 (65.2\%) for Soft-2D (Table~\ref{tab:heldout}). All three interfaces succeeded, so the gap is one of reliability rather than capability. Crucially, on the 30 primary scenes every interface selected the correct cube, yet completions diverged to 27, 21, and 19. The interface thus acts after selection: a correct direction label alone did not ensure the policy could translate it into approach, alignment, and lift.

Two controls identify why the population code helps. First, its advantage did not require trial-specific variation. Replacing each trial latent with the training-set mean of its decoded class (removing within-direction trial variation while keeping a distinct 128-dimensional code per direction) completed 83/102 grasps (81.4\%), matching trial-level Full-Latent (80/102, 78.4\%; Table~\ref{tab:packaging}). The stable, high-dimensional class structure itself is the useful condition.

Second, the input pathway matters as much as the representation. The same one-hot code completed 70/102 grasps (68.6\%) when appended to the robot state but only 55/102 (53.9\%) when sent through the dedicated-token pathway used by Full-Latent (Table~\ref{tab:packaging}). A 14.7-point swing from an otherwise identical code makes input placement a first-order engineering consideration, not an implementation detail.

The pattern across conditions points to population geometry as the effective scaffold. Hard-7 presents the seven directions as unrelated symbols; Soft-2D retains only their external coordinates; Full-Latent and the class prototype expose dense, direction-specific patterns learned from population activity. With only 207 demonstrations, that structure appears to organize the target-to-action mapping that a bare symbol cannot.

\begin{table}[t]
\centering
\caption{Grasp completion on all held-out scenes (count, \%).}
\label{tab:heldout}
\footnotesize
\begin{tabular}{lccc}
\toprule
Scenes & \full{} & \hard{} & \soft{} \\
\midrule
All 132 & 107 (81.1) & 91 (68.9) & 86 (65.2) \\
\bottomrule
\end{tabular}
\end{table}

\begin{table*}[t]
\centering
\caption{Intent-code controls on 102 additional new-layout scenes.}
\label{tab:packaging}
\footnotesize
\begin{tabular}{lcccc}
\toprule
Code & Signal & Dim. & Input & Success \\
\midrule
\full{} & Trial latent & 128 & Dedicated token & 78.4\% \\
Class prototype & Class mean latent & 128 & Dedicated token & 81.4\% \\
\hard{} & One-hot class & 7 & Appended to state & 68.6\% \\
Hard-7-token & One-hot class & 7 & Dedicated token & 53.9\% \\
\bottomrule
\end{tabular}
\end{table*}

\subsection{PMd's Advantage Emerges Only When Goal and Movement Diverge}
\label{sec:maze-pmd}
The third question is where in cortex a useful goal signal lives. Prior work separates PMd goal representation from M1 activity that tracks the executed movement; we asked whether that distinction survives as a behavioral difference. A dual-array maze recording from monkey J provides the test: barriers or distractors dissociated the initial hand direction from the final target in 79\% of trials. Both areas were read over the same movement-aligned window ($\pm500$ ms), so the neural source---not timing or channel count---distinguishes the two conditions.

PMd decoded goal direction more accurately than M1 (97.2\% vs.\ 90.1\%) and drove more completed grasps: 102/132 (77.3\%) versus 70/132 (53.0\%; Table~\ref{tab:decodability}). A conventional M1 readout from the first post-go second completed only 56/132 (42.4\%). Wrong-object grasps were rare across conditions, so the difference is in execution after a target was supplied, not in selecting the wrong object.

Two analyses show this is not simply an offline-accuracy effect. First, on the 114 scenes where both decoders identified the target correctly, the PMd-conditioned policy still completed 91 grasps (79.8\%) versus 63 (55.3\%) for M1: the same correct label produced different behavior depending on its source. Second, adding M1 to PMd raised completion only from 102/132 to 108/132 (81.8\%) while lifting decoding by 0.5 points, so more channels alone do not explain PMd's advantage.

\begin{figure*}[t]
\centering
\includegraphics[width=.98\textwidth]{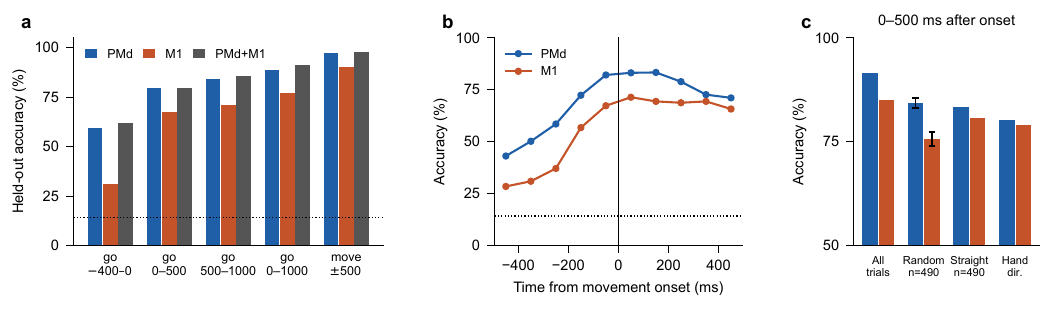}
\caption{Goal decoding depends on cortical source and readout window in a dual-array maze recording. (a) Held-out accuracy by window. (b) Decoding in 100-ms bins around movement onset. (c) Post-onset goal decoding for all trials, size-matched random trials, and straight reaches, followed by initial-hand-direction decoding. PMd retains more goal information than M1; the difference narrows when movement and goal align.}
\label{fig:areawin}
\end{figure*}

\begin{table*}[t]
\centering
\caption{Neural decoding accuracy, target identification, and grasp success by source and readout window (monkey J, 132 scenes).}
\label{tab:decodability}
\footnotesize
\setlength{\tabcolsep}{2.5pt}
\begin{tabular}{lccc}
\toprule
Source & Dec. & Target & Success \\
\midrule
M1, post-go 0--1000 ms & 77.0\% & 101/132 & 42.4\% \\
M1, move $\pm$500 ms & 90.1\% & 120/132 & 53.0\% \\
PMd, move $\pm$500 ms & 97.2\% & 126/132 & 77.3\% \\
PMd+M1, move $\pm$500 ms & 97.7\% & 129/132 & 81.8\% \\
\bottomrule
\end{tabular}
\par\smallskip
\parbox{.92\textwidth}{\footnotesize Dec., offline decoding accuracy; Target, correctly identified scene directions; Success, completed intended grasps.}
\end{table*}

The decisive control is the aligned case. On straight reaches, PMd and M1 decoded the goal similarly (83.3\% vs.\ 80.8\%); in the center-out task, M1 even decoded better (93.5\% vs.\ 87.0\%). PMd's advantage is therefore task-dependent: it appears when the policy must pursue a target independently of the initial movement and disappears when goal and movement align. That is the signature of a goal representation rather than a movement one, extending prior work~\cite{Cisek2005,Afshar2011} from decoding to embodied control. It also shows why the area with the highest offline score in a simple aligned task need not be the best intent source for complex embodied control.

\subsection{Preparatory Activity Already Carries a Usable Goal}
The final question is temporal: can a population state specify the goal before movement begins, and how does goal information change across preparation and execution? The first is tested in closed-loop episodes; the second, offline across sessions.

Across five readout windows in the dual-array recording, PMd decoded above M1 throughout. The gap was largest before the go cue (59.2\% vs.\ 31.4\%) and narrowed around movement onset (97.2\% vs.\ 90.1\%; Table~\ref{tab:windows}, Fig.~\ref{fig:areawin}); adding M1 changed decoding by at most 2.8 points. Once movement is underway both areas converge on the target, but before the go cue PMd already carried a distinctly clearer signal.

\begin{table}[t]
\centering
\caption{Direction accuracy by area and readout window (validation/test, \%).}
\label{tab:windows}
\footnotesize
\begin{tabular}{lccc}
\toprule
Window & PMd & M1 & PMd+M1 \\
\midrule
Go $-400$ to 0 ms & 68.9/59.2 & 41.7/31.4 & 70.0/62.0 \\
Go 0 to 500 ms & 86.7/79.8 & 69.2/67.6 & 85.0/79.6 \\
Go 500 to 1000 ms & 86.4/84.3 & 72.7/70.9 & 85.7/85.9 \\
Go 0 to 1000 ms & 93.4/88.5 & 81.1/77.0 & 92.0/91.3 \\
Movement $-500$ to $+500$ ms & 99.3/97.2 & 91.6/90.1 & 99.3/97.7 \\
\bottomrule
\end{tabular}
\end{table}

The closed-loop test of sufficiency read PMd+M1 during the 500 ms before the go cue in monkey M session B, when the target was known but no movement had started. This pre-go signal brought the robot to the intended cube in 29/30 scenes and completed 20 of those 29 grasps (69.0\%); substituting the intent reduced intended grasps to 1/30. The signal reliably specified which object to approach, even though the policy did not always finish the lift. Strikingly, the pre-go condition completed more grasps than the same session's post-cue Full-Latent readout (20/30 vs.\ 15/30; a post-cue PMd-only condition reached 17/30). A preparatory population state can therefore condition a complete grasp before movement begins.

The monkey M center-out sessions confirm the framework in an aligned-reach task: post-cue intent brought the robot to the intended object in 30/30 and 29/30 scenes, with 19/30 and 15/30 completed grasps, and substitution reduced correct selection to 2/30 scenes.

\begin{figure*}[t]
\centering
\includegraphics[width=.88\textwidth]{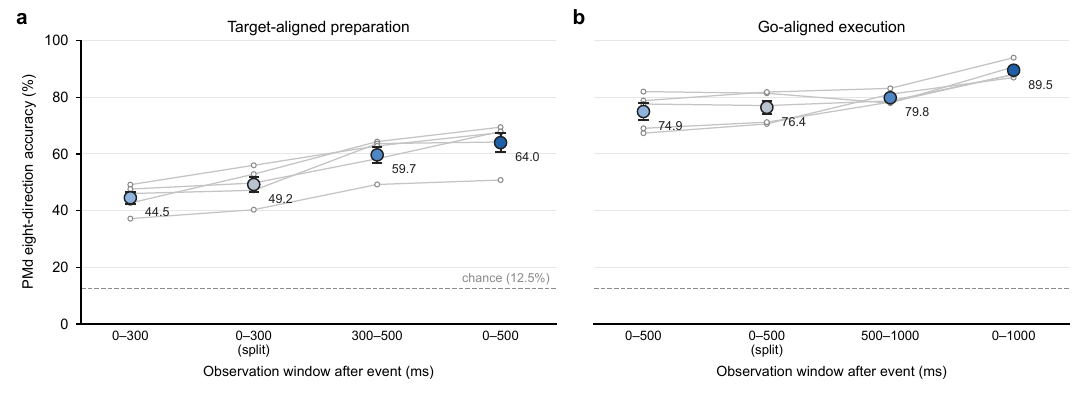}
\caption{Goal information accumulates in PMd across five sessions from monkey T. Accuracy is shown for early, late, dimension-matched split, and combined windows during preparation (a) and execution (b). Gray lines show individual sessions; markers show mean and standard error.}
\label{fig:timing}
\end{figure*}

Finally, offline analysis across five PMd sessions from monkey T shows the preparatory signal accumulates. Eight-direction decoding rose from 44.5\% in an early preparatory window to 64.0\% when late preparatory activity was added, above the 49.2\% dimension-matched early control, and reached 89.5\% after the go cue (Fig.~\ref{fig:timing}), with the gain positive in every session. Taken together, these results show a pre-go state can already serve as a task condition, and that its goal information grows across preparation.

\section{Discussion}
These experiments separate three properties of a cortical signal used as a task-level condition: whether it drives object choice, which representation best supports grasp completion, and where and when a goal is available. Counterfactual substitution addresses the first: replacing the neural trial while holding scene and policy fixed redirected object choice. The representation and source comparisons address the second: even after correct selection, policies differed in whether they completed the approach and lift, and the area that best served an aligned reach did not best serve a decoupled one. The timing results address the third: a pre-go PMd+M1 state already conditioned completed grasps, and offline decoding grew as preparatory activity accumulated. Decoding, selection, and completion therefore measure different parts of the interface.

The representation results show that both code structure and input pathway shape grasp performance. A class-mean latent matched trial-level Full-Latent without retaining trial-specific variation, identifying stable high-dimensional class structure as the useful signal. The pathway control was equally strong: the same one-hot code fell from 68.6\% to 53.9\% when moved from the state vector to a dedicated token, making input placement a first-order design variable rather than an implementation detail.

Held-out accuracy rules out crude label memorization, but it certifies only the seven-way class, not every latent dimension. That residual concern is answered by the class-prototype control: removing trial-specific variation left grasp completion intact, so the benefit rests on stable class geometry rather than trial-to-trial fluctuation.

The source comparison carries this distinction to behavior. In the maze, window-matched PMd decoding exceeded M1 (97.2\% vs.\ 90.1\%) and the PMd-conditioned policy completed far more intended grasps. A likely neural explanation is that M1 mixes target information with movement kinematics; those movement-specific components support direction decoding but give ACT trial-varying details that do not map onto the robot's action chunks, whereas PMd's more goal-selective activity gives a cleaner task condition~\cite{Cisek2005,Afshar2011,Cao2025PMd}. The aligned-task control confirms the interpretation: when goal and movement coincide, PMd's advantage disappears, and in center-out M1 even decodes better. PMd's value is therefore task-dependent, which is what a goal representation should show; offline decodability, by contrast, mainly reflects whether the label can be recovered.

Several limits constrain the conclusions. Neural recordings and robot demonstrations were independent and linked only by target direction; no participant controlled the simulated robot, and the neural signal received no robot feedback. The MC\_Maze decoder was evaluated on held-out trials from the same recording, so its accuracy establishes within-session trial generalization rather than transfer across sessions or animals. These gaps point to concrete next steps. Testing latent geometry on independently recorded sessions would show whether the learned code is stable across sessions, and repeating each policy across training seeds would quantify outcome variance. Extending the area and pre-go comparisons to more dual-array sessions and larger scene sets would test whether the effects replicate. Beyond the present setting---a simulated arm, three visible cubes, seven bearings, and spatial intent---future work should move to a physical robot and evaluate intent updates within an episode, continuous control, user workload, and human-in-the-loop use, alongside randomized codes that separate dimensionality from class geometry.

\section{Conclusion}
NI-ACT uses a brief motor-cortical population state as a persistent condition for a visuomotor policy. Counterfactual substitution shows that this condition directs object choice. Among the tested representations, Full-Latent produced the highest grasp completion, while class-prototype performance shows that trial-specific latent variation was not required. PMd was the more effective intent source when goal and initial movement were dissociated, and not when they aligned; a pre-go PMd+M1 state also conditioned complete grasps before movement began. Neural interface performance therefore depends on how a learned policy consumes the signal, not only on offline decoding accuracy.

\clearpage
\bibliographystyle{IEEEtran}
\bibliography{references}

@article{Hochberg2012,
  author  = {Hochberg, Leigh R. and Bacher, Daniel and Jarosiewicz, Beata and Masse, Nicolas Y. and Simeral, John D. and Vogel, Joern and Haddadin, Sami and Liu, Jie and Cash, Sydney S. and van der Smagt, Patrick and Donoghue, John P.},
  title   = {Reach and Grasp by People with Tetraplegia Using a Neurally Controlled Robotic Arm},
  journal = {Nature},
  year    = {2012},
  volume  = {485},
  pages   = {372--375}
}

@article{Collinger2013,
  author  = {Collinger, Jennifer L. and Wodlinger, Brian and Downey, John E. and Wang, Wei and Tyler-Kabara, Elizabeth C. and Weber, Douglas J. and McMorland, Angus J. C. and Velliste, Meel and Boninger, Michael L. and Schwartz, Andrew B.},
  title   = {High-Performance Neuroprosthetic Control by an Individual with Tetraplegia},
  journal = {The Lancet},
  year    = {2013},
  volume  = {381},
  pages   = {557--564}
}

@article{Wodlinger2015,
  author  = {Wodlinger, Brian and Downey, John E. and Tyler-Kabara, Elizabeth C. and Schwartz, Andrew B. and Boninger, Michael L. and Collinger, Jennifer L.},
  title   = {Ten-Dimensional Anthropomorphic Arm Control in a Human Brain--Machine Interface},
  journal = {Journal of Neural Engineering},
  year    = {2015},
  volume  = {12},
  pages   = {016011}
}

@article{Muelling2017,
  author  = {Muelling, Katharina and Venkatraman, Arun and Valois, Jean-Sebastien and Downey, John E. and Weiss, Jeffrey M. and Javdani, Shervin and Hebert, Martial and Schwartz, Andrew B. and Collinger, Jennifer L. and Bagnell, J. Andrew},
  title   = {Autonomy Infused Teleoperation with Application to Brain Computer Interface Controlled Manipulation},
  journal = {Autonomous Robots},
  year    = {2017},
  volume  = {41},
  number  = {6},
  pages   = {1401--1422}
}

@inproceedings{Zhang2023NOIR,
  title     = {{NOIR}: Neural Signal Operated Intelligent Robots for Everyday Activities},
  author    = {Zhang, Ruohan and Lee, Sharon and Hwang, Minjune and Hiranaka, Ayano and Wang, Chen and Ai, Wensi and Tan, Jin Jie Ryan and Gupta, Shreya and Hao, Yilun and Levine, Gabrael and Gao, Ruohan and Norcia, Anthony and Fei-Fei, Li and Wu, Jiajun},
  booktitle = {Proceedings of the 7th Conference on Robot Learning},
  pages     = {1737--1760},
  year      = {2023},
  volume    = {229},
  series    = {Proceedings of Machine Learning Research}
}

@article{Churchland2010,
  author  = {Churchland, Mark M. and Cunningham, John P. and Kaufman, Matthew T. and Ryu, Stephen I. and Shenoy, Krishna V.},
  title   = {Cortical Preparatory Activity: Representation of Movement or First Cog in a Dynamical Machine?},
  journal = {Neuron},
  year    = {2010},
  volume  = {68},
  pages   = {387--400}
}

@article{Churchland2012,
  author  = {Churchland, Mark M. and Cunningham, John P. and Kaufman, Matthew T. and Foster, Justin D. and Nuyujukian, Paul and Ryu, Stephen I. and Shenoy, Krishna V.},
  title   = {Neural Population Dynamics during Reaching},
  journal = {Nature},
  year    = {2012},
  volume  = {487},
  pages   = {51--56}
}

@inproceedings{Pei2021NLB,
  author    = {Pei, Felix and Ye, Joel and Zoltowski, David M. and Wu, Anqi and Chowdhury, Raeed H. and Sohn, Hansem and O'Doherty, Joseph E. and Shenoy, Krishna V. and Kaufman, Matthew T. and Churchland, Mark and Jazayeri, Mehrdad and Miller, Lee E. and Pillow, Jonathan and Park, Il Memming and Dyer, Eva L. and Pandarinath, Chethan},
  title     = {Neural Latents Benchmark '21: Evaluating Latent Variable Models of Neural Population Activity},
  booktitle = {Advances in Neural Information Processing Systems, Datasets and Benchmarks Track},
  year      = {2021}
}

@article{Pandarinath2018LFADS,
  author  = {Pandarinath, Chethan and O'Shea, Daniel J. and Collins, Jasmine and Jozefowicz, Rafal and Stavisky, Sergey D. and Kao, Jonathan C. and Trautmann, Eric M. and Kaufman, Matthew T. and Ryu, Stephen I. and Hochberg, Leigh R. and Henderson, Jaimie M. and Shenoy, Krishna V. and Abbott, Larry F. and Sussillo, David},
  title   = {Inferring Single-Trial Neural Population Dynamics Using Sequential Auto-Encoders},
  journal = {Nature Methods},
  year    = {2018},
  volume  = {15},
  pages   = {805--815}
}

@inproceedings{Zhao2023ACT,
  author    = {Zhao, Tony Z. and Kumar, Vikash and Levine, Sergey and Finn, Chelsea},
  title     = {Learning Fine-Grained Bimanual Manipulation with Low-Cost Hardware},
  booktitle = {Proceedings of Robotics: Science and Systems},
  year      = {2023}
}

@inproceedings{Chi2023DiffusionPolicy,
  author    = {Chi, Cheng and Feng, Siyuan and Du, Yilun and Xu, Zhenjia and Cousineau, Eric and Burchfiel, Benjamin and Song, Shuran},
  title     = {Diffusion Policy: Visuomotor Policy Learning via Action Diffusion},
  booktitle = {Proceedings of Robotics: Science and Systems},
  year      = {2023}
}

@inproceedings{Kim2025OpenVLA,
  author    = {Kim, Moo Jin and Pertsch, Karl and Karamcheti, Siddharth and Xiao, Ted and Balakrishna, Ashwin and Nair, Suraj and Rafailov, Rafael and Foster, Ethan P. and Sanketi, Pannag R. and Vuong, Quan and Kollar, Thomas and Burchfiel, Benjamin and Tedrake, Russ and Sadigh, Dorsa and Levine, Sergey and Liang, Percy and Finn, Chelsea},
  title     = {{OpenVLA}: An Open-Source Vision-Language-Action Model},
  booktitle = {Proceedings of the 8th Conference on Robot Learning},
  pages     = {2679--2713},
  year      = {2025},
  volume    = {270},
  series    = {Proceedings of Machine Learning Research}
}

@article{Musallam2004,
  author = {Musallam, S. and Corneil, B. D. and Greger, B. and Scherberger, H. and Andersen, R. A.},
  title = {Cognitive Control Signals for Neural Prosthetics},
  journal = {Science}, year = {2004}, volume = {305}, number = {5681}, pages = {258--262},
  doi = {10.1126/science.1097938}
}

@article{Aflalo2015,
  author = {Aflalo, Tyson and Kellis, Spencer and Klaes, Christian and Lee, Brian and Shi, Ying and Pejsa, Kelsie and Shanfield, Kathleen and Hayes-Jackson, Stephanie and Aisen, Mindy and Heck, Christi and Liu, Charles and Andersen, Richard A.},
  title = {Decoding Motor Imagery from the Posterior Parietal Cortex of a Tetraplegic Human},
  journal = {Science}, year = {2015}, volume = {348}, number = {6237}, pages = {906--910},
  doi = {10.1126/science.aaa5417}
}

@article{Cisek2005,
  author = {Cisek, Paul and Kalaska, John F.},
  title = {Neural Correlates of Reaching Decisions in Dorsal Premotor Cortex: Specification of Multiple Direction Choices and Final Selection of Action},
  journal = {Neuron}, year = {2005}, volume = {45}, number = {5}, pages = {801--814},
  doi = {10.1016/j.neuron.2005.01.027}
}

@article{Afshar2011,
  author = {Afshar, Afsheen and Santhanam, Gopal and Yu, Byron M. and Ryu, Stephen I. and Sahani, Maneesh and Shenoy, Krishna V.},
  title = {Single-Trial Neural Correlates of Arm Movement Preparation},
  journal = {Neuron}, year = {2011}, volume = {71}, number = {3}, pages = {555--564},
  doi = {10.1016/j.neuron.2011.05.047}
}

@article{Kaufman2014,
  author = {Kaufman, Matthew T. and Churchland, Mark M. and Ryu, Stephen I. and Shenoy, Krishna V.},
  title = {Cortical Activity in the Null Space: Permitting Preparation Without Movement},
  journal = {Nature Neuroscience}, year = {2014}, volume = {17}, number = {3}, pages = {440--448},
  doi = {10.1038/nn.3643}
}

@article{Elsayed2016,
  author = {Elsayed, Gamaleldin F. and Lara, Antonio H. and Kaufman, Matthew T. and Churchland, Mark M. and Cunningham, John P.},
  title = {Reorganization Between Preparatory and Movement Population Responses in Motor Cortex},
  journal = {Nature Communications}, year = {2016}, volume = {7}, pages = {13239},
  doi = {10.1038/ncomms13239}
}

@article{Downey2016,
  author = {Downey, John E. and Weiss, Jeffrey M. and Muelling, Katharina and Venkatraman, Arun and Valois, Jean-Sebastien and Hebert, Martial and Bagnell, J. Andrew and Schwartz, Andrew B. and Collinger, Jennifer L.},
  title = {Blending of Brain-Machine Interface and Vision-Guided Autonomous Robotics Improves Neuroprosthetic Arm Performance During Grasping},
  journal = {Journal of NeuroEngineering and Rehabilitation}, year = {2016}, volume = {13}, pages = {28},
  doi = {10.1186/s12984-016-0134-9}
}

@inproceedings{Javdani2015,
  author = {Javdani, Shervin and Srinivasa, Siddhartha and Bagnell, Andrew},
  title = {Shared Autonomy via Hindsight Optimization},
  booktitle = {Robotics: Science and Systems XI}, year = {2015},
  doi = {10.15607/rss.2015.xi.032}
}

@article{Dragan2013,
  author = {Dragan, Anca D. and Srinivasa, Siddhartha S.},
  title = {A Policy-Blending Formalism for Shared Control},
  journal = {The International Journal of Robotics Research}, year = {2013}, volume = {32}, number = {7}, pages = {790--805},
  doi = {10.1177/0278364913490324}
}

@article{Wu2006Kalman,
  author = {Wu, Wei and Gao, Yun and Bienenstock, Elie and Donoghue, John P. and Black, Michael J.},
  title = {Bayesian Population Decoding of Motor Cortical Activity Using a Kalman Filter},
  journal = {Neural Computation}, year = {2006}, volume = {18}, number = {1}, pages = {80--118},
  doi = {10.1162/089976606774841585}
}

@article{Gilja2012Refit,
  author = {Gilja, Vikash and Nuyujukian, Paul and Chestek, Cindy A. and Cunningham, John P. and Yu, Byron M. and Fan, Joline M. and Churchland, Mark M. and Kaufman, Matthew T. and Kao, Jonathan C. and Ryu, Stephen I. and Shenoy, Krishna V.},
  title = {A High-Performance Neural Prosthesis Enabled by Control Algorithm Design},
  journal = {Nature Neuroscience}, year = {2012}, volume = {15}, number = {12}, pages = {1752--1757},
  doi = {10.1038/nn.3265}
}

@article{Glaser2020Decoding,
  author = {Glaser, Joshua I. and Benjamin, Ari S. and Chowdhury, Raeed H. and Perich, Matthew G. and Miller, Lee E. and Kording, Konrad P.},
  title = {Machine Learning for Neural Decoding},
  journal = {eNeuro}, year = {2020}, volume = {7}, number = {4}, pages = {ENEURO.0506-19.2020},
  doi = {10.1523/ENEURO.0506-19.2020}
}

@article{Gallego2020,
  author = {Gallego, Juan A. and Perich, Matthew G. and Chowdhury, Raeed H. and Solla, Sara A. and Miller, Lee E.},
  title = {Long-Term Stability of Cortical Population Dynamics Underlying Consistent Behavior},
  journal = {Nature Neuroscience}, year = {2020}, volume = {23}, number = {2}, pages = {260--270},
  doi = {10.1038/s41593-019-0555-4}
}

@article{Orsborn2014,
  author = {Orsborn, Amy L. and Moorman, Helene G. and Overduin, Simon A. and Shanechi, Maryam M. and Dimitrov, Dragan F. and Carmena, Jose M.},
  title = {Closed-Loop Decoder Adaptation Shapes Neural Plasticity for Skillful Neuroprosthetic Control},
  journal = {Neuron}, year = {2014}, volume = {82}, number = {6}, pages = {1380--1393},
  doi = {10.1016/j.neuron.2014.04.048}
}

@article{AICopilot2025,
  author = {Lee, Johannes Y. and Lee, Sangjoon and Mishra, Abhishek and Yan, Xu and McMahan, Brandon and Gaisford, Brent and Kobashigawa, Charles and Qu, Mike and Xie, Chang and Kao, Jonathan C.},
  title = {Brain-Computer Interface Control with Artificial Intelligence Copilots},
  journal = {Nature Machine Intelligence}, year = {2025}, volume = {7}, number = {9}, pages = {1510--1523},
  doi = {10.1038/s42256-025-01090-y}
}

@article{Shared3D2023,
  author = {Zhou, Yajun and Yu, Tianyou and Gao, Wei and Huang, Weichen and Lu, Zilin and Huang, Qiyun and Li, Yuanqing},
  title = {Shared Three-Dimensional Robotic Arm Control Based on Asynchronous {BCI} and Computer Vision},
  journal = {IEEE Transactions on Neural Systems and Rehabilitation Engineering}, year = {2023}, volume = {31}, pages = {3163--3175},
  doi = {10.1109/tnsre.2023.3299350}
}

@article{ARShared2024,
  author = {Kokorin, Kirill and Zehra, Syeda R. and Mu, Jing and Yoo, Peter and Grayden, David B. and John, Sam E.},
  title = {Semi-Autonomous Continuous Robotic Arm Control Using an Augmented Reality Brain-Computer Interface},
  journal = {IEEE Transactions on Neural Systems and Rehabilitation Engineering}, year = {2024}, volume = {32}, pages = {4098--4108},
  doi = {10.1109/tnsre.2024.3500217}
}

@inproceedings{Black2025Pi0,
  author    = {Black, Kevin and others},
  title     = {$\pi_0$: A Vision-Language-Action Flow Model for General Robot Control},
  booktitle = {Proceedings of Robotics: Science and Systems (RSS)},
  year      = {2025},
  note      = {arXiv:2410.24164}
}

@inproceedings{Liu2025RDT,
  author    = {Liu, Songming and Wu, Lingxuan and Li, Bangguo and Tan, Hengkai and Chen, Huayu and Wang, Zhengyi and Xu, Ke and Su, Hang and Zhu, Jun},
  title     = {{RDT-1B}: a Diffusion Foundation Model for Bimanual Manipulation},
  booktitle = {International Conference on Learning Representations (ICLR)},
  year      = {2025},
  note      = {arXiv:2410.07864}
}

@misc{Bjorck2025GR00T,
  author       = {{NVIDIA} and Bjorck, Johan and others},
  title        = {{GR00T N1}: An Open Foundation Model for Generalist Humanoid Robots},
  howpublished = {arXiv:2503.14734},
  year         = {2025}
}

@article{Cao2025PMd,
  author  = {Cao, Sheng-Hao and Han, Xin-Yong and Zhao, Zhi-Ping and Gu, Jian-Wen and Jiang, Tian-Zi and Yu, Shan},
  title   = {Hand Position Fields of Neurons in the Premotor Cortex of Macaques during Natural Reaching},
  journal = {Nature Communications},
  year    = {2025},
  volume  = {16},
  number  = {1},
  pages   = {3489},
  doi     = {10.1038/s41467-025-58786-3}
}
\end{document}